\documentclass[11pt]{article}
\usepackage{booktabs}
\usepackage{amsmath}
\usepackage{multirow}
\usepackage{url}
\usepackage{microtype}
\usepackage[preprint]{acl}
\usepackage{amssymb} 
\usepackage{amsmath} 
\usepackage{times}
\usepackage{latexsym}
\usepackage[T1]{fontenc}

\usepackage[utf8]{inputenc}

\usepackage{microtype}

\usepackage{inconsolata}

\usepackage{algorithm}
\usepackage{algorithmic}
\usepackage{graphicx}
\usepackage{xcolor}
\usepackage[normalem]{ulem}
\NewDocumentCommand{\fk}{m m o}{%
  {\color{red}\sout{#1}}%
  \ {\color{blue}#2}%
  \IfValueT{#3}{%
    \ {\color{gray}\footnotesize[#3]}%
  }%
}
\newcommand{\fwa}{HyGRL}

\usepackage{caption}
\usepackage{enumitem}
\usepackage[most]{tcolorbox} 

\tcbset{
    promptbox/.style={
        enhanced,
        colback=white,
        colframe=black!75,
        fonttitle=\bfseries\sffamily,
        coltitle=white,
        attach boxed title to top left={xshift=5mm, yshift=-2mm},
        boxed title style={colback=black!75, sharp corners},
        title={#1},
        sharp corners,
        boxrule=0.8pt,
        breakable, 
        top=1.5em
    },
    casebox/.style={
        enhanced,
        colback=white,
        colframe=black,
        boxrule=0.8pt,
        sharp corners,
        parbox=false,
        before upper={\parindent1.5em}
    }
}

\title{HyGRL: Adaptive Hybrid Graph Reasoning for Multi-Entity Questions}

\author{
  Junyi Wang$^{1}$ \and
  Kaiyu Feng$^{1,*}$ \and
  Zixin Wang$^{1}$ \and
  Liaopeng Qiao$^{1}$ \and
  Ye Yuan$^{1}$ \\
  $^{1}$Beijing Institute of Technology \\
  $^{*}$Correspondence Author
}

\begin{document}
\maketitle
\begin{abstract}

Multi-entity compositional questions pose significant challenges to existing retrieval-augmented language models. Conventional methods fall into a dilemma: standard RAG lacks dynamic reasoning, traditional Graph-RAG is limited by structural sparsity, and LLM-constructed Graph-RAG incurs prohibitive costs. We propose \textbf{\fwa}, a unified framework that embeds unstructured text into structured knowledge graphs, creating a heterogeneous network for flexible evidence retrieval. Reasoning is formulated as adaptive structure induction, learned via a robust two-stage process: (1) imitation learning distills heuristic expert signals, and (2) reinforcement learning refines the policy using LLM-driven preference rewards.
Experiments demonstrate that {\fwa} effectively merges textual richness with structural knowledge, outperforming SOTA baselines in answer accuracy and reasoning fidelity while maintaining extremely low token costs and near real-time inference((code available at \url{https://github.com/wjywjy123/HyGRL}) .
\end{abstract}

\section{Introduction}
\label{sec:intro}

\begin{figure}[t]
\includegraphics[width=\columnwidth]{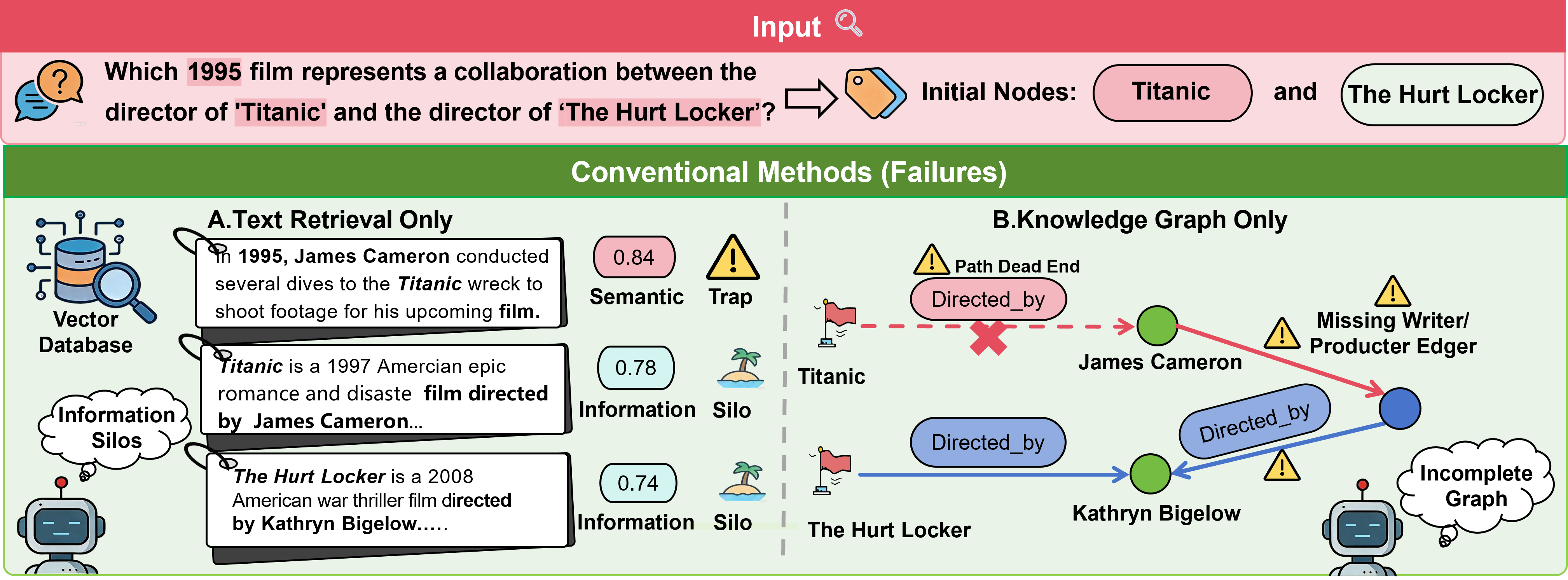}
  \caption{Text chunks or knowledge graphs alone are often insufficient to answer multi-entity compositional questions. }
  \label{fig:experiments}
\end{figure}

Retrieval-Augmented Generation (RAG) is crucial for mitigating factual hallucinations in Large Language Models (LLMs). While Standard RAG \cite{gao2023survey,lewis2020rag} relies on vector similarity to retrieve isolated text fragments, Graph-based RAG \cite{graphrag} explicitly models entity associations to provide structured reasoning paths.
Despite their widespread adoption, both approaches face insurmountable bottlenecks when processing \textbf{Multi-entity Compositional Questions}. These questions require models to perform multi-hop reasoning across multiple entities to find the answer (example in Figure \ref{fig:experiments}). 


To answer such questions, \textbf{(1) Standard RAG struggles with multi-hop reasoning} because dense retrieval tends to prioritize passages with high query similarity while missing distributed bridging evidence (Figure~\ref{fig:experiments}A). \textbf{(2) Existing Graph RAG methods remain constrained by KG structural sparsity.} As shown in Figure~\ref{fig:experiments}B, missing relations frequently interrupt reasoning paths. Our analysis on Freebase and Wikidata (Appendix~\ref{sec:appendix_sparsity}) further shows that even densely maintained KGs remain structurally incomplete for multi-hop reasoning: although Wikidata successfully covers 85.6\% of queried entities, only 37.6\% of query-answer pairs can be connected within three hops.

To address these bottlenecks, prior work has explored integrating text with graphs. However, methods leveraging LLMs to extract triples from text corpora (e.g., GraphRAG \cite{graphrag}, LightRAG \cite{lightrag}) incur high token costs and suffer from information loss when handling complex relations. Alternatively, approaches using multi-layer text indices (e.g., PathRAG \cite{pathrag}, HippoRAG \cite{hipporag, hipporag2}) treat text chunks merely as static retrieval endpoints rather than active reasoning elements, thereby preventing cross-modal transitions and limiting compositional flexibility.

However, improving graph construction alone is insufficient, as efficient reasoning over heterogeneous structures remains challenging. Existing retrieval methods often rely on fixed heuristic propagation (e.g., Personalized PageRank \cite{hipporag}), limiting their ability to learn query-specific reasoning behaviors. Agent-based frameworks \cite{tog, stride, bridgerag} improve adaptivity through iterative LLM-guided exploration, but incur substantial online token costs and latency. Reinforcement Learning (RL) offers a promising alternative, yet existing approaches either suffer from sparse rewards \cite{knowgpt} or require costly full-model optimization \cite{graph-r1}.

To address these limitations, we propose \fwa, a unified framework for multi-entity compositional reasoning. Instead of relying on either sparse symbolic graphs or isolated text retrieval alone, {\fwa} jointly leverages structured knowledge and unstructured textual evidence within a unified reasoning space. The framework further enables adaptive and efficient reasoning through lightweight policy learning, avoiding expensive online LLM-based search.


\noindent\textbf{Contributions:}

\noindent $\bullet$ 
We unify structured KGs and unstructured text by embedding text chunks as relay nodes within the graph, alleviating structural sparsity without costly triple extraction.

\noindent $\bullet$ 
We formulate reasoning as adaptive structure induction over heterogeneous graphs and train a lightweight policy via a two-stage paradigm combining heuristic imitation learning and LLM-based preference reinforcement learning.



\noindent $\bullet$ Experiments show that {\fwa}  achieves SOTA accuracy and fidelity, with computational efficiency and LLM robustness. 
\section{Related Work}

\subsection{Unimodal RAG: Text and KG}
Standard RAG relies on the semantic similarity of discrete text chunks but suffers from semantic isolation and the ``lost in the middle'' phenomenon. To mitigate this, methods like Beam Retrieval \cite{textBeam} expand the search space, while iterative strategies (\cite{iter-retgen, self-rag} dynamically refine retrieval. However, their frequent LLM interactions incur high token costs and latency. Although CompactRAG \cite{compactrag} restructures external knowledge offline to reduce online overhead, and RT-RAG \cite{rt-rag} decompose queries into reasoning trees, purely text-based methods still lack explicit structural reasoning. Conversely, KG-centric methods \cite{gcr, rog, tog} explicitly trace logical paths but heavily depend on graph completeness. They frequently fail when encountering sparse relations or missing entities, inevitably causing reasoning chains to break.

\subsection{Heterogeneous Text-Graph Construction}
To mitigate the structural sparsity of knowledge graphs, early graph neural network frameworks \cite{HGN, graft-net, pullnet} attempted to directly incorporate text chunks into the construction of heterogeneous graphs to bridge semantic associations. Recently, global text-graph indexing based on Large Language Models (LLMs) has emerged as a mainstream paradigm: GraphRAG \cite{graphrag} constructs hierarchical graph summaries to handle macro-level queries; pathRAG \cite{pathrag} specifically extracts and indexes key relational paths; and HippoRAG \cite{hipporag} draws inspiration from the human hippocampus to achieve associative memory retrieval. However, these methods collectively suffer from a heavy reliance on expensive offline LLM extraction and are highly prone to losing deep semantic information during the forced structurization process.

\subsection{Agent-based Graph RAG}
At the retrieval execution level, HippoRAG \cite{hipporag} employs a simple Personalized PageRank (PPR) random walk, while HGRAG utilizes hypergraph diffusion but relies on hard-coded rules. To enhance multi-hop accuracy, HippoRAG 2 \cite{hipporag2} introduces online LLM triple filtering, and ToG2 \cite{tog2} directly deploys the LLM as an agent to dynamically extend logical chains, which consequently incurs prohibitive token costs. KnowGPT \cite{knowgpt} attempts to employ reinforcement learning (RL) to train a small model for path filtering, yet it remains constrained by sparse rewards.

Recently, the field has shifted towards \textbf{End-to-End RL} frameworks (e.g., Graph-R1 \cite{graph-r1}, Graph-RFT \cite{graph-rft}, AutoGraph-R1 \cite{autograph-r1}, TeaRAG \cite{tearag}), which update full LLM parameters to jointly optimize retrieval and generation. Despite strong performance, these approaches suffer from heavy-parameter optimization, requiring expensive gradient back-propagation on billions of parameters and facing high feedback latency. In contrast, we propose a lightweight update paradigm: by freezing the LLM as a supervisor and training a small MLP policy network, we provide dense, millisecond-level feedback, ensuring robust convergence with zero-token routing during inference.

\begin{figure*}[t] 
  \centering
  \includegraphics[width=0.9\textwidth]{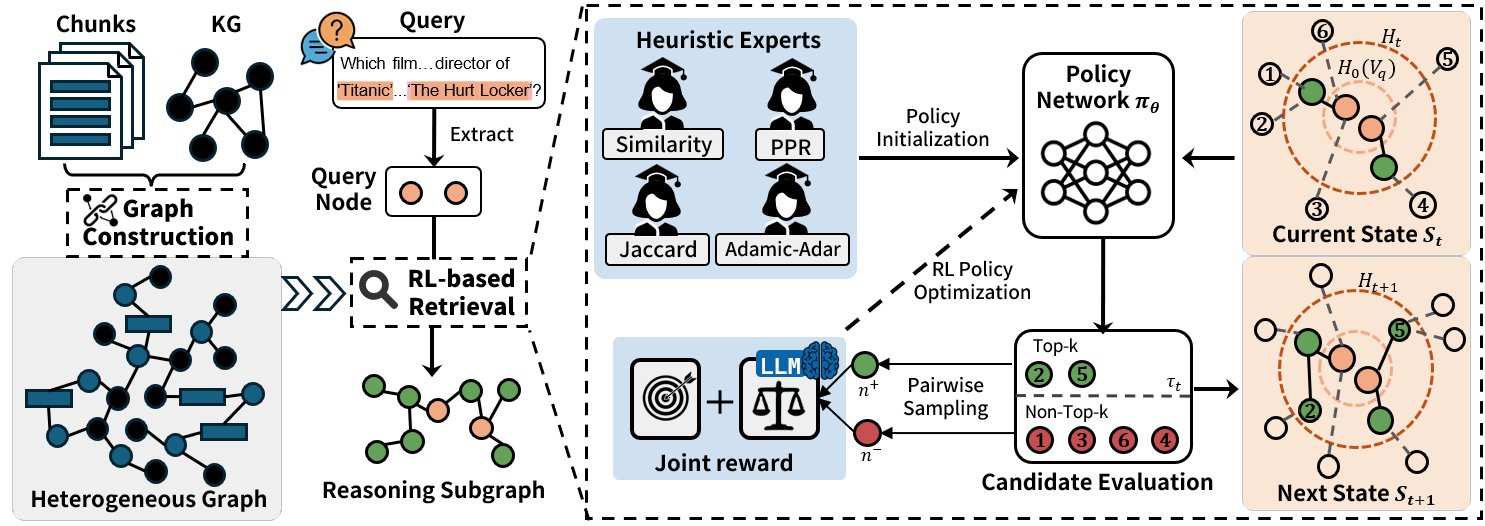}
  \caption{Overall architecture of {\fwa}. \textbf{Left}: Heterogeneous graph construction. \textbf{Center}: System input (query) and final reasoning output ($G_{sub}$). \textbf{Right (dashed box)}: The adaptive retrieval process, where current state $S^t$ is processed by policy $\pi_\theta$ to select top-$k$ nodes for expansion.  The MLP policy is initialized via heuristic experts and further refined by LLM-scored preference pairs (top-$k$ vs. non-top-$k$) and global ground-truth rewards.}
  \label{fig:wide_image}
\end{figure*}
\section{Problem Formulation}
As introduced in Section \ref{sec:intro}, 
multi-entity compositional questions require integrating evidence distributed across multiple entities, relations, and textual contexts. 
To formalize this task, we assume access to two complementary knowledge sources:
\noindent \textbf{(1) Unstructured Text Corpus} $\mathcal{T} = \{t_1, t_2, \dots, t_N\}$, which provides up-to-date and fine-grained contextual information.
\noindent\textbf{(2) Structured Knowledge Graph} $\mathcal{G} = (\mathcal{E}, \mathcal{R})$, which encodes entities and their relations in an explicit relational structure.

The task is to enable an LLM to generate a correct answer $A_{ans}$ to question $Q$ by conditioning on evidence retrieved from both $\mathcal{T}$ and $\mathcal{G}$. Instead of modifying the LLM, we focus on learning a reasoning and retrieval process that produces supporting evidence $E$ for the LLM. Formally, we maximize:
\begin{equation}
\hat{E} = \arg\max_{E} P(E \mid Q, \mathcal{T}, \mathcal{G}) ,
\end{equation}
which is then used with $Q$ as input to the LLM to generate $A_{ans}$.




\section{Methodology}

We propose \textbf{Hy}brid \textbf{G}raph reasoning with \textbf{R}einforcement \textbf{L}earning ({\fwa}), a three-stage framework: (1) constructing a heterogeneous graph unifying text chunks and a KG to merge textual freshness with relational structure;(2) formulating reasoning as adaptive structure induction over this graph, where the system incrementally expands from question entities to gather relevant evidence, initially guided by heuristic strategies; and (3) refining the reasoning policy via LLM preference feedback to align with downstream performance. 
The framework is shown in Figure~\ref{fig:wide_image}.

\subsection{Heterogeneous Graph Construction}\label{subsec:Graph_Construction}
Traditional LLM-based subject-predicate-object triplet extraction from text is costly, prone to hallucinations, and isolates text chunks from dynamic reasoning.
To overcome these limitations, we extract only entities and directly incorporate the raw text chunks into the graph, thereby constructing a heterogeneous graph $G=(V,R)$.
Specifically, we use an LLM to extract (Entity, Type) tuples from text chunks, linking them to KG nodes $\mathcal{V}$ via type-based disambiguation, constituting the entity set $V_e$ (the  prompt is in Appendix \ref{sec:appendix_prompts}).
Raw text chunks are then incorporated as distinct nodes $V_c$ and bidirectionally connected to their mentioned entities ($V_e \leftrightarrow V_c$).

To avoid online computation bottlenecks over the massive global KG, we construct a text corpora specific heterogeneous graph offline. We first process all text chunks within the corpus, since they contain up-to-date contextual information. We extract all entities mentioned in these text chunks and then connect them through shortest paths over the KG, forming a comprehensive offline index that preserves relational evidence for the entire dataset.

\subsection{RL-based Reasoning}

We model reasoning over the heterogeneous graph as a Markov Decision Process (MDP). Unlike prior methods assuming fixed reasoning structures (e.g., paths or trees), multi-entity questions require flexible, convergent reasoning. Thus, we employ reinforcement learning to train a policy network that guides an adaptive beam search for evidence retrieval.

\noindent\textbf{State and Action Space.}
Reasoning starts from entities mentioned in the question (seed nodes, denoted as $V_Q$) and incrementally adds nodes to form an ordered sequence. Due to the heterogeneity of the graph, the added nodes comprise three structures: seed entities, unstructured text chunks, and explicit relational triples.To limit state size, we retain only the most recent $L$ nodes, forming a sliding-window state $N_s = \langle n_1, \dots, n_L\rangle$($|N_s| \le L$) At each step, multiple top-$k$ neighbors (beam width $k$) can be added to expand the sequence.

To ensure generalization to unseen questions and avoid coupling the state space to specific training questions, we design a relevance-based representation that is invariant to specific text realizations. Specifically, the state is represented by a relevance vector $S \in \mathbb{R}^{L}$, where $S[i]$ is the textual similarity between node $n_i$ and the question $Q$. This representation captures how relevant each selected node is to the question, without encoding raw textual features.By avoiding computationally expensive online dense encoders, this design choice facilitates the framework's near real-time inference speed and minimal memory footprint.

The action space consists of unselected neighbors of the current state nodes. For a candidate action $a$, we construct an action vector $A \in \mathbb{R}^{L+1}$, where $A[i]$ encodes the similarity between $a$ and node $n_i$, and $A[L+1]$ includes the PPR score. This allows the policy to assess whether selecting $a$ introduces new evidence beyond the current reasoning context.

\noindent\textbf{Policy Network.}
The reasoning policy $\pi_\theta$ is parameterized by a lightweight multi-layer perceptron (MLP). For a state vector $S$ and vector $A$, the network predicts the selection probability $P(a)$ via a Sigmoid activation:
$P(a) = \pi_\theta(S, A) = \sigma(\mathrm{MLP}([S \parallel A]))$.
This design enables efficient evaluation of candidate actions while remaining agnostic to specific node identities or graph schemas.

\noindent\textbf{Adaptive Search Procedure.}
Guided by the policy network, the agent explores the graph via a modified beam search (Algorithm \ref{alg:graph_search}), using a confidence threshold $\tau$ to filter expansion noise.
At each step, we aggregate unvisited neighbors $\mathcal{N}(n_i)$ from all nodes in the current sequence $N_s$ into a candidate set $C$, retaining only those with $P(a) \ge \tau$ to form $C_{valid}$.
The top-$k$ highest-probability nodes from $C_{valid}$ are appended to $N_s$ and marked as visited in $V_{vis}$.
Search terminates upon reaching $D_{max}$, state length $L$, or exhausting valid candidates. The final reasoning subgraph is induced as $G_{sub} = G[N_s]$.

\begin{algorithm}[ht]
\caption{Adaptive Beam Search}
\label{alg:graph_search}
\small 
\begin{algorithmic}[1]
\renewcommand{\algorithmicrequire}{\textbf{Input:}}
\renewcommand{\algorithmicensure}{\textbf{Output:}}
\REQUIRE Graph $G$, Seeds $V_Q$, Policy $\pi_\theta$, Beam width $k$ \\
\textit{Params}: Threshold $\tau$, State limit $L$, Max steps $D_{max}$
\ENSURE Subgraph $G_{sub}$

\STATE $N_s \leftarrow V_Q$, \ $V_{vis} \leftarrow V_Q$, \ $d \leftarrow 0$
\WHILE{$d < D_{max}$ \AND $|N_s| < L$}
    \STATE $C_{vld} \leftarrow \{a \in \mathcal{N}(n_i) \setminus V_{vis} \mid n_i \in N_s, P(a) \ge \tau\}$
    \STATE \textbf{if} $C_{vld} = \emptyset$ \textbf{then} \textbf{break}
    
    \STATE $C_{top} \leftarrow \mathop{\arg\max}\limits^{(k)}_{a \in C_{vld}} P(a)$
    \STATE $N_s \leftarrow N_s \oplus C_{top}$, \ $V_{vis} \leftarrow V_{vis} \cup C_{top}$, \ $d \leftarrow d + 1$
\ENDWHILE
\RETURN $G_{sub} \leftarrow G[N_s]$
\end{algorithmic}
\end{algorithm}

\subsection{Policy Initialization}\label{subsec:IL}
Training the reasoning policy requires a carefully designed supervision signal, which is non-trivial in our setting. A straightforward approach is to define the reward based on whether the retrieved evidence enables a large language model to correctly answer a multi-entity compositional question.
However, such outcome-based supervision leads to extremely sparse rewards, since meaningful feedback is only available after the entire reasoning process and final answer generation. This makes effective credit assignment for intermediate expansion decisions difficult and hampers learning efficiency.

To mitigate this challenge and enable efficient policy initialization, we introduce a set of complementary  heuristic expansion strategies and adopt imitation learning to distill their behavior into an initial reasoning policy. We next present a set of heuristic methods for imitation learning.

\noindent\textbf{Context-aware Semantic Matching ($m_{ce}$)}: Computes the semantic interaction between $a$ and the context $(Q, N_s)$ using a cross-encoder:
$m_{ce}(a, Q, N_s) = \text{Enc}(Q \oplus N_s, \text{Text}(a))$.

\noindent\textbf{Multi-source Personalized PageRank ($m_{ppr}$)}: Using a restart vector $\mathbf{r}$ centered on $V_Q$, this score reflects $a$'s global structural importance relative to the query:
$m_{ppr}(a) = \alpha \sum_{v \in \mathcal{N}_{in}(a)} \frac{m_{ppr}(v)}{|\mathcal{N}_{out}(v)|} + (1-\alpha) r_a$,
where $v$ represents an in-neighbor of $a$, and $\mathcal{N}_{in}(\cdot)$ and $\mathcal{N}_{out}(\cdot)$ denote the sets of incoming and outgoing neighbors, respectively.

\noindent\textbf{Path-aware Adamic-Adar Index ($m_{aa}$)}: Evaluates structural closeness between the candidate node $a$ and the current node $v_{cur}$ by weighting shared neighbors,where $v_{cur} \in N_s$ is the node from which the candidate $a$ is being expanded (i.e., $a \in \mathcal{N}(v_{cur})$):
$m_{aa}(a) = \sum_{z \in \mathcal{N}(a) \cap \mathcal{N}(v_{cur})} \frac{1}{\log |\mathcal{N}(z)|}.$

\noindent\textbf{Jaccard Coefficient ($m_{jac}$)}: Measures the neighbor overlap ratio between $a$ and $v_{cur}$:
$m_{jac}(a) = \frac{|\mathcal{N}(a) \cap \mathcal{N}(v_{cur})|}{|\mathcal{N}(a) \cup \mathcal{N}(v_{cur})|}$.

With the four heuristic methods, we next introduce a hybrid fusion mechanism to combine them.

\smallskip\noindent\textbf{Hybrid Fusion Mechanism}: To leverage the complementary strengths of the four heuristics, we introduce a learnable parameter set $\Theta = \{\theta_{ce}, \theta_{ppr}, \theta_{aa}, \theta_{jac}\}$ to adaptively fuse their signals. These parameters are normalized via softmax to obtain weights $w_k = \exp(\theta_k) / \sum_{j} \exp(\theta_j)$, and the aggregated score for a candidate node $a$ is computed as the weighted sum $S_{s}(a) = \sum w_k(\Theta) \cdot m_k(a)$. Specifically, $\Theta$ is calibrated on a small random subset of the training set and subsequently frozen (see Appendix \ref{sec:appendix_gold_neighbors} for details).

\smallskip\noindent\textbf{Optimization}:
Using $S_s(a)$ as a teacher, we distill the policy via supervised learning. $\pi_\theta(S, A)$ is trained to minimize the binary cross-entropy loss $\mathcal{L}_{IL}(\theta) = \text{BCE}(y_a, \pi_\theta(S, A))$, where $y_a=1$ for high-confidence heuristic nodes.This process transfers the prior knowledge inherent in heuristics to the parameterized policy.

\subsection{Policy Optimization}\label{subsec:RL_ft}

While imitation learning ensures efficient initialization, static heuristics may be misaligned with the downstream LLM's reasoning priorities.
To address this, we further refine the policy via reinforcement learning, using step-wise LLM preference feedback to align the agent's decisions with the LLM's judgment.

\smallskip\noindent\textbf{Local Pairwise Reward.}
Obtaining absolute scores from an LLM can be unreliable, as predicted values may not be comparable across actions.
We thus adopt a pairwise approach: the LLM acts as a ``preference mentor'' to determine which of two candidate actions is superior given state $S$. This formulation ensures robust supervision while reducing sensitivity to absolute score scale.

The LLM evaluates action pairs based on logical consistency, information completeness, and structural clarity(see Appendix \ref{sec:appendix_prompts} for the pairwise preference prompt). To limit computational cost, we sample candidate node pairs $(n^+, n^-)$ from the action space $\mathcal{A}_t$ at each state. Specifically, $n^+$ is drawn from the policy's Top-$K$ candidates, and $n^-$ from the remaining neighbors. Let $A^+$ and $A^-$ be their corresponding action vectors.
We stably update the policy network via a margin ranking loss on the pre-activation logits $f_{\theta}(S,A) = \text{MLP}([S \parallel A])$:

\begin{equation}
\small
    \mathcal{L}_{MR}(\theta) = \sum \max\left(0, \xi - \left( f_\theta(S, A^+) - f_\theta(S, A^-) \right) \right)
\end{equation}

where $\xi$ is the margin. This aligns the policy with the LLM's logical preferences without requiring a separate reward model.


\smallskip\noindent\textbf{Joint Optimization.}
Beyond step-wise feedback, we ensure global reasoning quality via Self-Critical Sequence Training (SCST), which evaluates the entire retrieval trajectory upon completion. Specifically, an LLM evaluates the retrieved subgraph's sufficiency for deducing the ground-truth answer via a strict binary prompt (Appendix \ref{sec:appendix_prompts}). For each training query, the model generates two complete trajectories: a sampled rollout and a greedy rollout, yielding global outcome rewards $r_{sample}$ and $r_{greedy}$ respectively, where the baseline $r_{greedy}$ is derived by deterministically selecting the most probable action at each step.The SCST loss then uses this greedy baseline to optimize the model, encouraging sampled actions that achieve higher overall rewards:
\begin{equation}
\small
    \mathcal{L}_{SCST}(\theta) = - (r_{sample} - \gamma  \cdot r_{greedy}) \times \sum \log \pi_\theta(S, A),
\end{equation}
where $\gamma$ is a scaling factor controlling the self-critical baseline.
To balance long-distance global planning with high-frequency local guidance, we integrate these signals into a unified objective $\mathcal{L}_{Total} = \mathcal{L}_{SCST} + \lambda \cdot \mathcal{L}_{MR}$, where $\lambda$ governs the trade-off between global reasoning quality and local supervision.
\subsection{Answer Generation}
Post-retrieval, the model yields a Reasoning Subgraph covering multiple complementary paths. This subgraph is verbalized into natural language evidence to augment the LLM's input, enabling synergy between structured reasoning and generative capabilities.


\section{Experiments}
\label{sec:experiments}

We investigate four research questions: (RQ1) the overall performance of {\fwa} compared to state-of-the-art baselines, (RQ2) the contributions of  components, (RQ3) computational efficiency, (RQ4) reasoning fidelity ,(RQ5) robustness to LLM hallucinations and (RQ6) parameter sensitivity.

\subsection{Experimental Setup}
\label{subsuc:Experimental Setup}
\noindent\textbf{Datasets.}
We evaluate our framework on three challenging multi-hop QA benchmark datasets: 2WikiMultiHopQA (2Wiki)~\cite{2wiki}, HotpotQA~\cite{hotpotqa}, and MuSiQue~\cite{musique}. The text corpus utilizes the official Wikipedia paragraph dumps provided by each benchmark. To focus on multi-entity compositional reasoning, we filter the query sets, retaining only samples with at least two seed entities successfully linked to the graph. All datasets are strictly partitioned into mutually isolated training, development, and test sets. Partitioning details and dataset statistics are presented in Appendix C.

\smallskip\noindent\textbf{Baselines.}
To ensure absolute empirical rigor and fairness, all baselines are re-evaluated under a fully identical experimental environment and the same filtered subsets, utilizing completely unified evaluation metrics with all operational costs standardized. We compare {\fwa}  against the following retrieval paradigms and structural constraints:

\noindent\textbf{(1) Standard RAG:} We include \textit{GPT-4o (Zero-shot)} ,\textit{textRAG} (standard top-K semantic retrieval) and \textit{TextBeamRAG}~\cite{textBeam}, a strong text-only baseline performing end-to-end beam search over chunks without KG constraints.

\noindent\textbf{(2) Graph-based RAG:} We include \textit{Microsoft GraphRAG}~
\cite{graphrag} for community-based global indexing; \textit{PathRAG}~\cite{pathrag} for semantic path retrieval; \textit{HippoRAG}~\cite{hipporag} for PPR-based search over heterogeneous graphs; \textit{Kg2RAG}~\cite{kg2rag} for external KG-guided chunk expansion; and \textit{HippoRAG2-hybrid}, modified based on HippoRAG2~\cite{hipporag2} to simultaneously yield both text chunks and entity nodes for output alignment.

\noindent\textbf{(3) Variants:} \textit{HyGRL-Heuristic} is evaluated to validate the necessity of the learned routing policy compared to heuristic expert scoring.

\smallskip\noindent\textbf{Metrics.}
 We evaluate both answer quality and efficiency. For quality, we report \textbf{Exact Match} rate (EM), the percentage of predictions exactly matching the normalized ground truth, and \textbf{F1 Score}, the word-level harmonic mean of precision and recall computed following the standard evaluation protocols of multi-hop QA benchmarks. For efficiency, we measure token consumption, latency, and memory usage.

\smallskip\noindent\textbf{Implementation Details.}
Freebase (40M entities) serves as the KG data source. All experiments are conducted on a server equipped with an NVIDIA A100 (80GB) GPU and 128GB RAM, utilizing PyTorch and NetworkX. The global KG is managed as a static backend; during entity linking retrieval, we employ SQLite for exact matching and GPU-accelerated FAISS for semantic vector search , completing offline index construction in about 3 hours. Inference latency for both HyGRL and all baselines is uniformly measured on a per-query basis. Peak system RAM and inference VRAM consumption are dynamically monitored via the psutil library and torch.cuda, respectively.
To construct the input features, the system uses a lightweight Cross-Encoder to compute the pairwise textual similarity scores. These scalar scores are then concatenated to form the (2L+1)-dimensional state-action vectors (where zero-padding is applied if $|N_s| < L$), which are processed by a [512, 256, 128] MLP policy network.For Imitation Learning (Sec.~\ref{subsec:IL}), the MLP is trained using AdamW ($1 \times 10^{-3}$) with a ReduceLROnPlateau scheduler. 
For RL fine-tuning (Sec.~\ref{subsec:RL_ft}), we employ Adam ($5 \times 10^{-4}$) and MarginRankingLoss (1.0 margin). Both stages use a batch size of 256, and GPT-4o is used uniformly for final answer generation.

\subsection{Results and Analysis}

\subsubsection{Main Results (RQ1)}
As shown in Table~\ref{tab:main_results}, {\fwa} achieves the best overall average performance and secures the highest EM across all datasets. On the structured 2Wiki dataset, \textit{Kg2RAG} and \textit{GraphRAG} yield higher token-level F1, but their static expansion or global summarization triggers indiscriminate aggregation, introducing semantic noise that degrades EM. \textit{TextBeamRAG} lacks constraints from explicit entity relationships, rendering it susceptible to semantic drift during multi-hop iterations. {\fwa} utilizes reinforcement learning for adaptive routing and logical pruning to suppress error accumulation and noise during the retrieval process.

\begin{table*}[t]
\centering
\small
\renewcommand{\arraystretch}{1.3}
\begin{tabular}{llcccccccc}
\hline
 &  & \multicolumn{2}{c}{\textbf{2Wiki}} & \multicolumn{2}{c}{\textbf{HotpotQA}} & \multicolumn{2}{c}{\textbf{MuSiQue}} & \multicolumn{2}{c}{\textbf{Average}} \\
\cline{3-4} \cline{5-6} \cline{7-8}\cline{9-10}
\textbf{Group} & \textbf{Method}  & \textbf{EM} & \textbf{F1} & \textbf{EM} & \textbf{F1} & \textbf{EM} & \textbf{F1} & \textbf{EM} & \textbf{F1}\\
\hline
\multirow{3}{*}{standard RAG} & GPT-4o (Zero-shot) & 46.25 & 70.16& 30.74 & 43.38& 10.40 & 19.55& 29.13 & 44.36\\
 & textRAG & 62.92 & 66.78 & 34.73 & 47.45 & 12.42 & 20.98 & 36.69 & 45.07 \\
 & TextBeamRAG & 71.25 & 75.86 & 44.57 & 54.27 & 16.00 & 24.91 & 43.94 & 51.68 \\
\hline
\multirow{5}{*}{Graph-based RAG} & Microsoft GraphRAG & 70.41 & \underline{84.29} & 51.44 & 60.71 & 19.30 & 28.56 & 47.05 & 57.85 \\
 & PathRAG & 74.40 & 75.20 & 51.84 & 62.54 & 20.34& 32.60 & 48.86& 56.78 \\
 & HippoRAG & \underline{79.16} & 83.80 & 53.48 & 65.16 & 25.04& 32.90 & 52.56& 60.62 \\
 & Kg2RAG & 76.40 & \textbf{85.41} & 53.79 & 64.49 & 25.34 & 34.55 & 51.84 & \underline{61.48} \\
 & HippoRAG2-hybrid & 78.63 & 79.21 & 57.67 & 68.23 & \underline{26.60} & \underline{36.25} & \underline{54.30} & 61.23 \\
\hline
\multirow{1}{*}{Variants} & {\fwa}-Heuristic & 77.24 & 77.85 & \underline{59.79} & \underline{69.87} & 23.75& 32.50 & 53.59& 60.07 \\
\hline
 & \textbf{{\fwa} (Ours)} & \textbf{79.19}& 79.85 & \textbf{61.52} & \textbf{72.06} & \textbf{28.01}& \textbf{39.70} & \textbf{56.24} & \textbf{63.87} \\
\hline
\end{tabular}
\caption{Comparative performance of {\fwa} against baselines.}
\label{tab:main_results}
\end{table*}

\begin{table}[ht]
\small
    \centering
    \begin{tabular}{lc}
    \hline
    \textbf{Variant} & \textbf{Avg. F1 Score} \\
    \hline
        {\fwa} (Full) & \textbf{63.87}\\
        w/o Text Nodes & 51.90 \\
        w/o KG Completion & 45.07 \\
        Replace RL with BFS & 60.12 \\
        \hline
    \end{tabular}
      \caption{Ablation study.}
    \label{tab:ablation}
\end{table}
\subsubsection{Ablation Study (RQ2)}

We analyze the impact of different graph modalities. The ``w/o Text Nodes'' variant runs retrieval on a graph without text chunks, while ``w/o KG Completion'' effectively degenerates into a standard RAG framework. As shown in Table~\ref{tab:ablation}, ``w/o KG Completion'' suffers a more severe performance drop, highlighting the importance of dynamic graph search. Replacing the trained policy with a random BFS achieves an F1 score of 60.12\%, demonstrating the success of our heterogeneous graph construction. Finally, the full model incorporating the two-stage training strategy achieves the highest performance (63.87\% F1), validating that our retrieval policy provides necessary intelligent guidance.

\begin{figure}[t]
\includegraphics[width=\columnwidth]{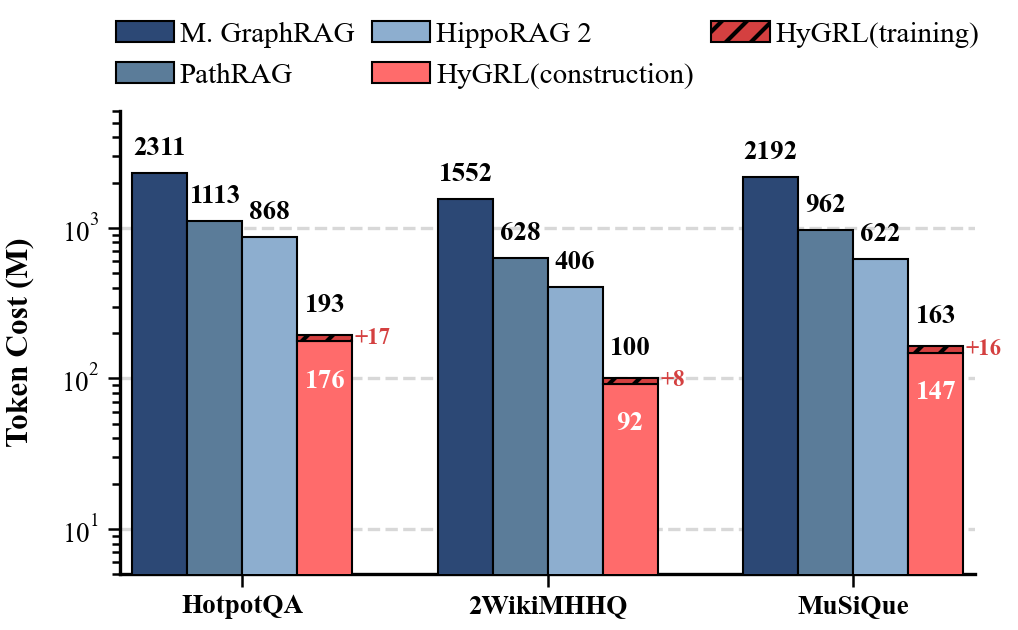}
  \caption{Comparison of Total Token Costs. For \textbf{{\fwa}}, the cost is decomposed into index construction (solid base) and offline training (shaded top).}
  \label{fig:token}
\end{figure}

\subsubsection{Efficiency and Deployment Costs (RQ3)}\label{subsec:cost}



We evaluate deployment costs by comparing LLM token consumption (Figure~\ref{fig:token}). During graph index construction, Microsoft GraphRAG consumes over 2 billion tokens using LLMs for triplet extraction and community summarization; PathRAG and HippoRAG also incur high costs due to their reliance on OpenIE. By contrast, {\fwa} achieves a 90\% reduction in construction costs via entity-centric extraction. Although {\fwa} involves offline training (comprising $N_{gold}$ annotation, preference sampling, and SCST verification), its cost is negligible compared to the massive construction overhead, as evidenced by the tiny red shaded areas in Figure~\ref{fig:token}. Ultimately, by internalizing retrieval logic within a lightweight MLP agent, {\fwa} achieves "zero-token retrieval", offering superior scalability over methods requiring iterative LLM calls \cite{tog, iter-retgen}.

Regarding efficiency and deployment feasibility (Table~\ref{tab:efficiency}), we evaluate the average inference latency and memory consumption on HotpotQA. Compared to a pure LLM (1.15s), \textbf{{\fwa} adds a mere 0.67s retrieval overhead, achieving a near real-time latency of 1.82s.} This speed is comparable to standard RAG and other graph-based methods. It avoids the latency associated with text-based methods requiring iterative retrieval (e.g., textBeamRAG at 2.25s), as well as the expensive delays of HippoRAG2-hybrid (2.91s), which requires online LLM filtering.

Furthermore, training the lightweight agent requires only $\sim$2 GB of VRAM since it solely updates the MLP; inference VRAM is exceptionally low at $\sim$475 MB, nearly on par with standard text retrieval. While \textbf{{\fwa}} requires higher system RAM (35.95 GB) than baselines, this is a deliberate trade-off to eliminate disk I/O latency. By keeping the heterogeneous graph entirely resident in memory, the system guarantees a stable sub-two-second response time.

\begin{table}[ht]
\small
    \centering
    \begin{tabular}{lcc}
    \hline
    \textbf{Method} & 
    \begin{tabular}{@{}c@{}}\textbf{Avg. Inference} \\ \textbf{Latency (s)}\end{tabular} & 
    \begin{tabular}{@{}c@{}}\textbf{System RAM} \\ \textbf{(GB)}\end{tabular} \\
    \hline
    LLM& 1.15& - \\
 textRAG& 1.93& 12.23\\
 TextBeamRAG& 2.25& 12.24\\
    PathRAG& 1.95& 9.28\\
    HippoRAG& 1.89& 16.26\\
 Kg2RAG& 1.75& 15.95\\
 HippoRAG2-hybrid& 2.91& 17.03\\
    \textbf{{\fwa} }& \textbf{1.82}& \textbf{35.95}\\
    \hline
    \end{tabular}
    \caption{Efficiency comparison on HotpotQA. 
    }
    \label{tab:efficiency}
\end{table}

\subsubsection{Evaluating Reasoning Fidelity (RQ4)}
\label{subsubsec:reasoning_fidelity}
To verify the reliability of the model's reasoning evidence, we evaluate on the 2WikiMultiHopQA dataset, which contains both textual and logical chain evidence. We uniformly parse the ground truth and retrieved evidence into text chunks and logical triples (triple outputs for pure-text baselines are recorded as N/A). By calculating precision and recall for exact matches, we derive three F1 metrics: SF-F1 and E-F1 independently measure the retrieval accuracy of pure text and structured triples, respectively; while UE-F1 computes the comprehensive F1 score on the mixed graph-text set to holistically assess the overall evidence coverage of heterogeneous retrieval. For baselines, we select textBeamRAG, pathRAG, and HippoRAG2-hybrid, which represent advanced iterative text retrieval, path retrieval with text as leaf nodes, and LLM-filtered hybrid retrieval paradigms, respectively.

As shown in Table~\ref{tab:fidelity_results}, \textbf{HyGRL} significantly outperforms the baselines in overall retrieval performance. Specifically, textBeamRAG suffers from semantic drift and lacks structural output (E-F1 is N/A); whereas HippoRAG2-hybrid achieves the highest E-F1 (21.9\%) via LLM triple filtering, it struggles to effectively align logic with textual semantics. In contrast, HyGRL's RL agent actively performs logical pruning and cross-modal transitions, achieving the optimal SF-F1 (44.7\%) and the highest overall UE-F1 (39.9\%). This demonstrates that heterogeneous graph systems must rely on intelligent routing policies to achieve perfect synergy between structure and text. Furthermore, the case studies in Appendix~\ref{sec:appendix_cases} qualitatively confirm its capability to precisely pinpoint evidence.

\begin{table}[ht]
    \centering
    \resizebox{\columnwidth}{!}{%
    \begin{tabular}{lccc}
    \hline
    \textbf{Method} & 
    \textbf{SF-F1 (\%)} & 
    \textbf{E-F1 (\%)} & 
    \textbf{UE-F1 (\%)} \\
    \hline
    textBeamRAG & 19.7 & N/A & 17.0 \\
    pathRAG & 21.4 & 16.6 & 19.9 \\
    HippoRAG2-hybrid & 30.4 & \textbf{21.9} & 28.6 \\
    \textbf{HyGRL (Ours)} & \textbf{44.7} & 17.1 & \textbf{39.9} \\
    \hline
    \end{tabular}%
    }
    \caption{Reasoning fidelity evaluation on 2Wiki.}
    \label{tab:fidelity_results}
\end{table}

\subsubsection{LLM Mentor Robustness  (RQ5)}
\label{subsubsec:robustness}


To evaluate the impact of LLM hallucinations, we inject controlled noise during the RL phase on HotpotQA by artificially flipping the pairwise preference labels.
As shown in Table~\ref{tab:noise_robustness}, \textbf{{\fwa}} demonstrates graceful degradation. At a 15\% noise rate, the F1 score drops by only \textbf{0.9}, and the \textbf{decline remains limited} even at 30\%; a significant drop occurs only at $p=50\%$ (random guessing). This robustness confirms that combining imitation learning with our objective joint-reward mechanism effectively mitigates inherent LLM biases, avoiding over-reliance on pure LLM rewards.
\begin{table}[ht]
\small
    \centering
    \begin{tabular}{lcccc}
    \hline
    \textbf{Noise Ratio ($p$)} & \textbf{0\% }& \textbf{15\%} & \textbf{30\%} & \textbf{50\%} \\
    \hline
    F1 Score& 50.34 & 49.44 & 48.28 & 46.25 \\
    $\Delta$ Drop & - & -0.90 & -2.06& -4.09 \\
    \hline
    \end{tabular}
    \caption{Performance under varying levels of injected preference noise. }
    \label{tab:noise_robustness}
\end{table}

To empirically validate the LLM mentor's supervision quality, three domain experts blindly assessed 100 preference pairs sampled from HotpotQA (retained after prompt filtering). The LLM achieves an \textbf{88.0\% alignment} with expert consensus. This confirms our filtering mechanism successfully discards ambiguous instances, yielding a highly reliable reward signal.

\subsubsection{Parameter Sensitivity Analysis (RQ6)}
\label{subsubsec:Parameter_Sensitivity}

\begin{figure}[t]
\includegraphics[width=\columnwidth]{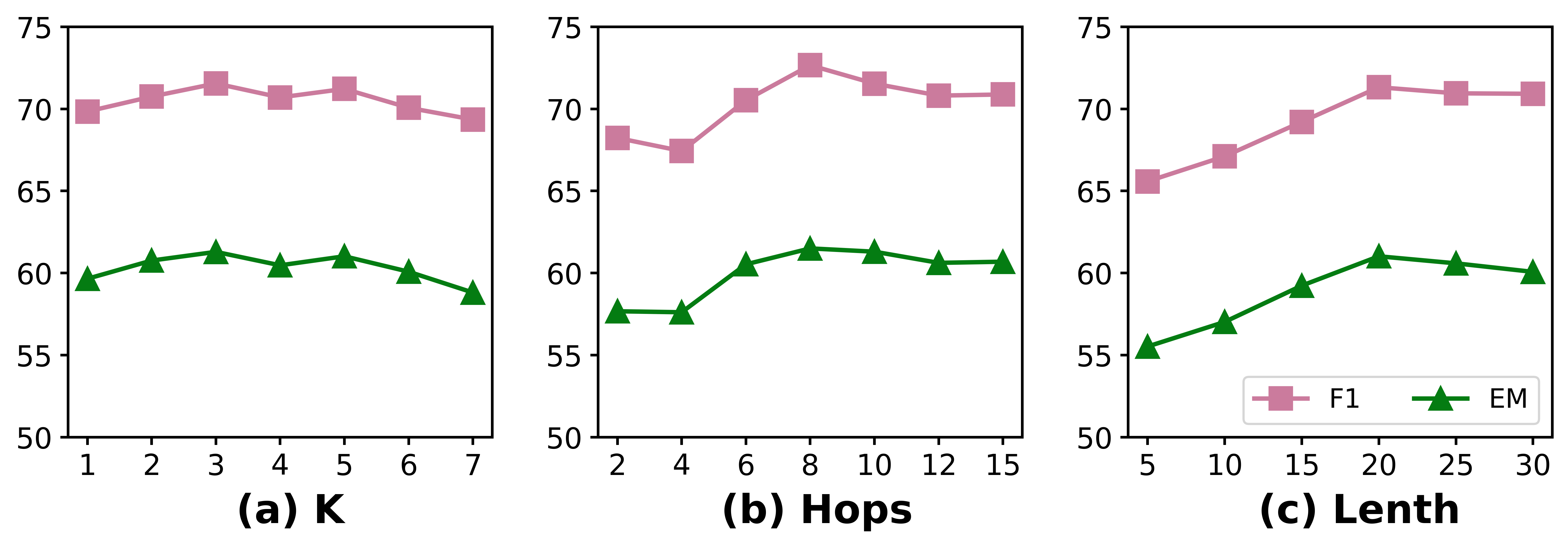}
  \caption{Sensitivity w.r.t. Beam Width ($K$), Search Depth (Hops), and Max Nodes Numbers (L).}
  \label{fig:sensitivity}
\end{figure}

We evaluate {\fwa}'s sensitivity to key hyperparameters (Figure~\ref{fig:sensitivity}).
Increasing \textbf{Beam Width ($K$)} improves performance up to an optimal value of 3, beyond which metrics decline.
For \textbf{Search Depth (Hops)}, performance experiences a significant peak at a depth of 8 before declining. 
Finally, for \textbf{Max Nodes Numbers ($L$)}, performance peaks at a length of 20 and deteriorates beyond 25.

\section{Conclusion}
In this paper, we introduced HyGRL, a hybrid graph reasoning framework that integrates text and KGs with reinforcement learning to guide LLM-based multi-entity compositional reasoning. By combining heuristic initialization and LLM preference feedback, HyGRL learns adaptive reasoning policies that outperform state-of-the-art baselines while remaining token-efficient, demonstrating the effectiveness of hybrid knowledge integration for complex reasoning tasks.

\section{Limitations}
Our framework derives preference-based supervision from LLMs, making performance dependent on the model’s consistency and domain expertise. Although we mitigate noise via pairwise discrimination and self-critical baselines, the resulting policy remains an approximation of the LLM’s implicit reasoning.
\bibliography{custom}
\clearpage
\appendix
\onecolumn 

\section{Empirical Analysis on KG Sparsity and Generalizability}
\label{sec:appendix_sparsity}

To empirically validate that structural sparsity is a universal bottleneck across different knowledge graphs (KGs) and to assess the generalizability of our heterogeneous graph framework, we conducted a pilot analysis. We randomly sampled 200 complex multi-hop queries from the HotpotQA and Musique datasets to compare three graph configurations: (1) \textbf{Pure Freebase (Subgraph)}, representing a relatively sparse structural baseline; (2) \textbf{Pure Wikidata}, queried via official APIs, representing a dense and up-to-date global KG environment; and (3) \textbf{HyGRL (Ours)}, which augments the sparse Freebase baseline by integrating unstructured text chunks as independent nodes. We measured the \textit{Entity Linking Rate} (the proportion of query start entities and answer end entities successfully mapped to the graph) and the \textit{Answer Connectivity Rate} (the percentage of queries with a valid path within 3 hops, calculated strictly on samples where both the start and end entities exist in the graph).

\begin{table}[ht]
\centering
\small
\caption{Empirical comparison of multi-hop structural connectivity across KGs.}
\label{tab:appendix_kg_comparison}
\begin{tabular}{lcc}
\toprule
\textbf{Graph Configuration} & \textbf{Entity Linking Rate (\%)} & \textbf{Answer Connectivity Rate (\%)} \\
\midrule
Pure Freebase & 77.0 & 24.1 \\
Pure Wikidata & \textbf{85.6} & 37.6 \\
\textbf{HyGRL (Ours)}& 77.0 & \textbf{69.7} \\
\bottomrule
\end{tabular}
\end{table}

As shown in Table~\ref{tab:appendix_kg_comparison}, although Wikidata's active maintenance boosts the Entity Linking Rate to 85.6\%, its Answer Connectivity Rate remains severely constrained at 37.6\%. This compellingly proves that physical missing edges are an inherent limitation of static KGs rather than a specific artifact of legacy databases. In contrast, HyGRL dramatically elevates the Answer Connectivity Rate to 69.7\%. Unstructured text chunks effectively act as \textit{Semantic Bridges}, successfully complementing missing relational links without requiring a perfectly dense KG backend. Freebase provides readily downloadable, complete graph structure data that facilitates local construction. \textbf{Furthermore, the empirical results fully confirm that our framework is inherently KG-agnostic, exerting a strong compensatory effect and robustness against incomplete structured metadata.}

\section{Extraction of Gold Neighbors for Policy Initialization}\label{sec:appendix_gold_neighbors}
As noted in Section 4.3, the heuristic fusion parameter set $\Theta$ is calibrated on a small random subset of the training data. To achieve this, we construct a set of high-quality "gold" transitions (denoted as $N_{gold}$) to serve as the ground-truth targets for calibration. Given that the four scalar parameters in $\Theta$ converge rapidly, this sampling strategy effectively minimizes LLM API costs and computational overhead. To enhance factual accuracy, we first map dataset supporting facts to the graph via exact or semantic matching, forming a target node set $V_{target}$ of text chunks and entities. The core objective is to explore reasoning paths starting from the seed nodes $V_Q$ extracted from the initial question to $V_{target}$, and to strictly filter out topologically connected but logically irrelevant noise, thereby extracting high-quality nodes as $N_{gold}$. To prune irrelevant noise, a reverse BFS from $V_{target}$ builds a distance map recording the shortest topological path to the target for each traversed node. This search process features a dual termination condition: it halts early upon reaching the seed nodes $V_Q$, or forcibly terminates when a predefined maximum number of hops is reached. During forward expansion, we retain only candidate nodes that decrease the distance to $V_{target}$ according to this map, effectively pruning off-target nodes from the candidate set $\mathcal{C}$.

Furthermore, an LLM is introduced to conduct point-wise logical verification on the candidate set $\mathcal{C}$ (the exact prompt template is provided in Appendix \ref{sec:appendix_prompts}). Validated nodes form $N_{gold}$ ($y^*_a=1$) to supervise the calibration loss, which is defined as the binary cross-entropy loss $\mathcal{L}_{1}(\Theta) = \text{BCE}(y^*_a, \sigma(S_s(a)))$, where $\sigma(\cdot)$ represents the standard Sigmoid function and $S_s(a)$ is the aggregated teacher score derived in Section 4.3. Once $\mathcal{L}_{1}(\Theta)$ is minimized over this sampled subset, the calibrated parameter set $\Theta$ is completely frozen and directly applied as invariant scaling constants to compute $S_s(a)$ over the remaining full training set during the subsequent imitation learning phase.

\section{ Dataset Details}
\label{sec:appendix_datasets}
Regarding the filtering mechanism mentioned in Section \ref{subsuc:Experimental Setup}, we implement a uniform entity extraction, linking, and filtering protocol across all dataset splits (including the training, development, and test sets). For HotpotQA, as the official test ground-truth is unreleased, we randomly partition the publicly available development set into two mutually exclusive, equal halves to serve as our validation and evaluation test sets, respectively. For 2Wiki and MuSiQue, standard randomized partitioning is applied to establish standalone splits. The core statistical profiles of the filtered and partitioned evaluation datasets are summarized in Table \ref{tab:dataset_stats}. Throughout our pipeline, the development sets are strictly restricted to optimal checkpoint selection and downstream hyperparameter tuning (e.g., verifying the beam width $k$ and state sequence limit $L$ analyzed in Section \ref{subsubsec:Parameter_Sensitivity}), ensuring the evaluation test sets remain completely untouched prior to final inference.
\begin{table}[htbp]
\label{dataset_statics}
\centering
\caption{Statistics of the filtered evaluation datasets.}
\label{tab:dataset_stats}
\begin{tabular}{lccc}
\hline
\textbf{Metric} & \textbf{MuSiQue} & \textbf{HotpotQA} & \textbf{2Wiki} \\ \hline
Questions                    & 500              & 976               & 1200            \\
Num. of Passages             & 6177             & 9395              & 8096            \\
Avg. Passage Length (Words)  & 74.81            & 86.28             & 57.90           \\ \hline
\end{tabular}
\end{table}
\section{Case Studies}
\label{sec:appendix_cases}

In this appendix, we present the comprehensive reasoning paths for four additional multi-hop queries from our experiments. To accommodate the complexity of the reasoning chains and the length of the supporting evidence, each case is presented below.

\subsection*{Case Study 1: Disambiguation via Textual Bridge (United States)}

To intuitively demonstrate the efficacy of {\fwa} in integrating structured knowledge with unstructured text, we examine the reasoning chain for the question: \textit{``In what county is Shady Grove, in the state where the Standard Glass and Paint Company building is located?''} The process begins by identifying the anchor entity, \textbf{Standard Glass and Paint Company}. While the KG provides basic topological connections, the agent enhances this by retrieving a text chunk which explicitly states the building is ``located in downtown Des Moines, Iowa.'' This \textit{textual bridge} effectively grounds the search context to the state of \textbf{Iowa}. Leveraging this constraint, the agent utilizes simple triple relations (e.g., \texttt{location.containedby}) to filter potential candidates for ``Shady Grove,'' correctly identifying the specific instance located within Iowa. Finally, the agent traverses to \textbf{Buchanan County} and corroborates the finding with a description chunk confirming Shady Grove as a community in that county. This instance exemplifies how our method synergizes the structural guidance of triples with the rich semantic details of text chunks to pinpoint the ``golden evidence'' amidst ambiguous paths.

\begin{tcolorbox}[casebox]
    \centering
    \vspace{0.2cm}
    \textbf{\large Case Study 1: Step-by-Step Reasoning Path}
    \vspace{0.2cm} \hrule \vspace{0.2cm}
    
    \raggedright
    \textbf{Question:} In what county is Shady Grove, in the state where the Standard Glass and Paint Company building is located? \\
    \textbf{Ground Truth:} Buchanan County \quad | \quad \textbf{Predicted:} Buchanan County
    
    \vspace{0.2cm} \hrule \vspace{0.2cm}
    
    \textbf{Reasoning Path:}
    \begin{itemize}[leftmargin=0.5cm]
        \item \textbf{Phase 1: Anchor Entity Identification \& Textual Bridge}
        \begin{itemize}
            \item \textbf{Reference Entity:} Standard Glass and Paint Company
            \item \textbf{Text Evidence (Contextual):} ``...located in downtown Des Moines, Iowa.''
            \item \textbf{Constraint Inferred:} The target state is grounded to \textbf{Iowa}.
        \end{itemize}
        
        \vspace{0.2cm}
        
        \item \textbf{Phase 2: Candidate Filtering via Structural Guidance}
        \begin{itemize}
            \item \textbf{Target Entity Candidates:} Shady Grove (Ambiguous)
            \item \textbf{Triple:} \texttt{[Shady Grove] $\xrightarrow{location.containedby}$ [Iowa]}
            \item \textbf{Model Inference Note:} Leveraging the state constraint established in Phase 1, the agent utilizes simple triple relations to filter potential candidates for ``Shady Grove,'' correctly identifying the specific instance located within Iowa.
        \end{itemize}

        \vspace{0.2cm}

        \item \textbf{Phase 3: Final Resolution}
        \begin{itemize}
            \item \textbf{Target Entity:} Buchanan County
            \item \textbf{Text Evidence (Contextual):} A description chunk corroborating the finding by confirming Shady Grove as a community within Buchanan County.
        \end{itemize}
    \end{itemize}
    \vspace{0.1cm}
\end{tcolorbox}
\captionof{figure}{Detailed breakdown of the reasoning chain for Shady Grove, illustrating the synergy between textual bridges and structural filters.}
\label{fig:case_shady_grove}

\vspace{2em}

The following sections detail the complete reasoning frameworks for three additional diverse scenarios, further demonstrating the model's robust multi-hop capabilities across different domains.

\vspace{2em}

\begin{tcolorbox}[casebox]
    \centering
    \vspace{0.2cm}
    \textbf{\large Case Study 2: Recursive Administrative Hierarchy (Burkina Faso)}
    \vspace{0.2cm} \hrule \vspace{0.2cm}
    
    \raggedright
    \textbf{Question:} What department includes the village of Douré, in the country where the Moussodougou Department is also found? \\
    \textbf{Ground Truth:} Zimtenga Department \quad | \quad \textbf{Predicted:} Zimtenga Department
    
    \vspace{0.2cm} \hrule \vspace{0.2cm}
    
    \textbf{Reasoning Path:}
    \begin{itemize}[leftmargin=0.5cm]
        \item \textbf{Phase 1: Identify Core Entity \& Determine Country Scope}
        \begin{itemize}
            \item \textbf{Reference Entity:} Moussodougou Department
            \item \textbf{Triple:} \texttt{[Moussodougou Department] $\xrightarrow{containedby}$ [Burkina Faso]}
            \item \textbf{Text Evidence (Full):} ``Moussodougou is a department or commune of Comoé Province in southern Burkina Faso. Its capital lies at the town of Moussodougou. According to the 1996 census the department has a total population of 9,407.''
        \end{itemize}
        
        \vspace{0.2cm}
        
        \item \textbf{Phase 2: Locate Target Entity within Context}
        \begin{itemize}
            \item \textbf{Target Entity:} Douré
            \item \textbf{Constraint:} Entity must be located within \textbf{Burkina Faso}
            \item \textbf{Triple:} \texttt{[Douré] $\xrightarrow{containedby}$ [Zimtenga Department]}
            \item \textbf{Text Evidence (Full):} ``Douré, Burkina Faso is a village in the Zimtenga Department of Bam Province in northern-central Burkina Faso. It has a population of 466.''
        \end{itemize}
    \end{itemize}
    \vspace{0.1cm}
\end{tcolorbox}
\captionof{figure}{Reasoning chain for disambiguating village locations using administrative hierarchy constraints.}
\label{fig:case_douree}

\vspace{2em}

\begin{tcolorbox}[casebox]
    \centering
    \vspace{0.2cm}
    \textbf{\large Case Study 3: Etymological Reasoning (Norway)}
    \vspace{0.2cm} \hrule \vspace{0.2cm}
    
    \raggedright
    \textbf{Question:} What is the country where Nissedal is located named after? \\
    \textbf{Ground Truth:} north \quad | \quad \textbf{Predicted:} north
    
    \vspace{0.2cm} \hrule \vspace{0.2cm}
    
    \textbf{Reasoning Path:}
    \begin{itemize}[leftmargin=0.5cm]
        \item \textbf{Phase 1: Geographic Positioning (Village to Country)}
        \begin{itemize}
            \item \textbf{Core Entity:} Nissedal
            \item \textbf{Triple:} \texttt{[Nissedal] $\xrightarrow{containedby}$ [Norway]}
            \item \textbf{Text Evidence (Full):} ``Tveitsund is a village in Nissedal municipality, Norway. The urban area Tveitsund, which consists of Tveitsund and Treungen, has a population of 361.''
        \end{itemize}
        
        \vspace{0.2cm}
        
        \item \textbf{Phase 2: Etymological Origin Retrieval}
        \begin{itemize}
            \item \textbf{Target Entity:} Norway
            \item \textbf{Triple:} \texttt{[Norway] $\xrightarrow{etymology\_or\_namesake}$ [north]}
            \item \textbf{Text Evidence (Contextual):} ``Norway has a total area of and a population of 5,312,300 (as of August 2018). The country shares a long eastern border with Sweden (1,619 km or 1,006 mi long). Norway is bordered by Finland and Russia to the north-east, and the Skagerrak strait to the south, with Denmark on the other side... Even during polar night in the north, temperatures above freezing are commonplace on the coastline.''
            \item \textbf{Model Inference Note:} While the retrieved paragraph describes the geography and climate dominated by ``northern'' features, the model leverages the specific knowledge relation \texttt{common.topic.etymology\_or\_namesake} to infer the semantic origin ``north'' (Norðrvegr).
        \end{itemize}
    \end{itemize}
    \vspace{0.1cm}
\end{tcolorbox}
\captionof{figure}{Reasoning chain connecting a specific location to the etymological origin of its country.}
\label{fig:case_norway}

\vspace{2em}

\begin{tcolorbox}[casebox]
    \centering
    \vspace{0.2cm}
    \textbf{\large Case Study 4: Cross-Domain Attribute Retrieval (Uganda)}
    \vspace{0.2cm} \hrule \vspace{0.2cm}
    
    \raggedright
    \textbf{Question:} Who is the current Chief Justice of the country where the Achwa river is found? \\
    \textbf{Ground Truth:} Bart Magunda Katureebe \quad | \quad \textbf{Predicted:} Bart Magunda Katureebe
    
    \vspace{0.2cm} \hrule \vspace{0.2cm}
    
    \textbf{Reasoning Path:}
    \begin{itemize}[leftmargin=0.5cm]
        \item \textbf{Phase 1: Geographical Anchoring (River to Country)}
        \begin{itemize}
            \item \textbf{Core Entity:} Achwa River
            \item \textbf{Triple:} \texttt{[Achwa River] $\xrightarrow{countries\_flows\_through}$ [Uganda]}
            \item \textbf{Text Evidence (Full):} ``The Achwa River is a river of Uganda in eastern Africa. It flows through the northern central part of the country, draining much of Uganda's northern plateau and northeastern highlands, before crossing the border into South Sudan where it joins the White Nile. In South Sudan it is known as the Aswa River.''
        \end{itemize}
        
        \vspace{0.2cm}
        
        \item \textbf{Phase 2: Political Attribute Retrieval (Country to Official)}
        \begin{itemize}
            \item \textbf{Target Entity:} Uganda
            \item \textbf{Target Relation:} \texttt{government.governmental\_jurisdiction.chief\_justice}
            \item \textbf{Triple:} \texttt{[Uganda] $\xrightarrow{chief\_justice}$ [Bart Magunda Katureebe]}
            \item \textbf{Text Evidence (Full):} ``Bart Magunda Katureebe is a Ugandan judge and the Chief Justice of Uganda. He was appointed to that position on 5 March 2015. Before that, he was a justice of the Supreme Court of Uganda.''
        \end{itemize}
    \end{itemize}
    \vspace{0.1cm}
\end{tcolorbox}
\captionof{figure}{Reasoning chain traversing from physical geography to political hierarchy.}
\label{fig:case_uganda}

\section{Prompt Templates}
\label{sec:appendix_prompts}

This section provides the full text of the instructional prompts used in our framework.

\begin{tcolorbox}[promptbox={Entity Extraction Prompt}]
\begin{lstlisting}
Please perform fine-grained entity recognition and classification from the text below. Follow these guidelines rigorously:

Concrete Instance Requirement: Only extract explicitly named instances (no categories/theories/domains)
Exclusion Threshold: Discard all terms failing ANY of:
Proper noun capitalization OR
Unique identifier presence (model numbers/patent IDs) OR
Event timestamp specification


Entity Typing Framework (8 mutually exclusive categories):

1. people
- Scope: Individual human beings (real/fictional) OR human-centric roles requiring agency
- Must: Full names/titles with uniquely identifiable references
- Examples: "Stephen Hawking", "Harry Potter"
- Excluded Cases: Groups ("research team"), honorary titles without person reference ("MVP award")

2. organization
- Scope: Collective entities with operational structure (corporate/govt/non-profit)
- Examples: "World Health Organization", "Special Mixture No. 8", "Qihua Primary School"
- Excluded Cases: Cannot be represented as single human actor

3. culture  
- Scope: Creative works, symbolic systems & intellectual artifacts
- Examples: "Ice Age"(film), "The History of the Counter Intelligence Corp"(book)
- Key differentiator: Human-created abstract constructs
- Excluded Cases:Theoretical concepts ("game theory"),Unnamed manuscripts ("economic behavior analysis" x)

4. location
- Scope: Spatially definable regions/structures (physical/virtual)
- Examples: "Asotin"(city), "Facebook Metaverse", "Inti Punku"(place of interest)
- Inclusion: Both permanent ("Pacific Ocean") and temporary ("Pop-up store")

5. science 
- Scope: Natural/formal science entities & fundamental principles
- Examples: "(20097) 1994 UL2"(astronomy), protein body(biology)
- Excluded Cases: Academic disciplines ("biology" ),Generic concepts ("nuclear strategies" ),Applied implementations belong to Products

6. sports  
- Scope: Institutionalized competitive activities & infrastructure
- Examples: "FIFA World Cup", "Madison Square Garden", "Decathlon"
- Critical check: Must contain competitive element

7. products  
- Scope: Market-traded items/services (physical/digital/conceptual)
- Examples: "iPhone 15", "USS Blue Jay (AMc-23)"(ship),"LTspice"(software)
- Boundary rule: Must have commercialization potential
- Excluded Cases:  Generic services ("insurance" x) -> "Prudential Whole Life Policy" v

8. event
- Scope: Temporal occurrences with defined chronology
- Examples: "Chernobyl Disaster", "2024 Solar Eclipse", "Black Friday Sale"
- Mandatory: Documented start/end (implicit or explicit)
- Excluded Cases: Isolated time references ("2024","1997"x )

Validation Protocol:
1. Reject ambiguous entities ("company", "vehicle","1992") without proper noun specification
2. Discard metaphorical references ("the birth of AI") without concrete temporal boundaries
3. Flag borderline cases with confidence scores
4. Standalone time expressions such as '1992'

Please format response as follow with exact surface mentions:
entity1,TYPE1;entity2,TYPE2;....
If no relevant entities can be extracted, simply output 'None'.
Remove duplicate entities, retaining a single occurrence.

Text to analyze: 
[Input Text Placeholder]
\end{lstlisting}
\end{tcolorbox}

\vspace{1cm}

\begin{tcolorbox}[promptbox={Gold Neighbor Verification Prompt}]
\begin{lstlisting}
Task Description

You are evaluating a single retrieved candidate neighbor given the current reasoning queue and the known target answer.

The queue consists of text snippets and/or triples that have already been selected and should be treated as current working assumptions in the reasoning process. They define the current partial reasoning structure and direction, which may be incomplete or provisional, but must not be overturned without explicit support.

Determine whether adding this candidate to the queue constitutes a necessary and logically coherent step toward reaching the target answer.

Evaluation Rules (Apply IN ORDER, do NOT skip steps)

You must evaluate the candidate strictly in the following order. If the candidate fails an earlier rule, it should be immediately rejected (output 0) even if it looks acceptable under later rules.

Execution Priority Summary

Rule 1 filters out invalid candidates.
Rule 2 is the primary criterion.
Rule 3 is a secondary check applied only after Rule 2 is satisfied.
To be accepted as a Gold Neighbor, the candidate must pass all three rules.

Rule 1 - Context Consistency & Answer Reachability (Hard Gate)

Evaluate the candidate under the semantic and structural context defined by the current queue. The queue defines the current reasoning context and direction, which may be partial or provisional.

Reject the candidate immediately if it relies on unsupported free association unrelated to the current context, or if it arbitrarily overturns the current reasoning direction without justification. Do not accept candidates solely based on surface semantic similarity or keyword overlap.

Rule 2 - Missing Information Completion (Primary Criterion)

First determine:
1. what information is required to connect the question to the target answer,
2. what is already present in the current queue,
3. and what specific information is still missing to complete a plausible reasoning path to the target answer.

Missing information may include, but is not limited to:
A key entity or concept,
A necessary intermediate entity or role,
A relation, causal link, or dependency,
A factual constraint or evidence (e.g., authorship, time, location, identity, attribution),
A bridging fact that connects the current reasoning context to the target answer.

Such information may be introduced explicitly (via entities or relations) or implicitly (via text snippets that provide required facts or constraints).

Accept the candidate ONLY if it directly fills one clearly identifiable missing component or makes it clearly reachable from the current queue. You should be able to clearly name the missing information, as a concrete component of the answer path, that this candidate helps to fill.

Rule 3 - Structural Progress (Secondary)

Assuming the candidate passes Rules 1 and 2, verify its impact on the reasoning structure. The transition must make the reasoning path toward the target answer more explicit, more directed, and less ambiguous, by:
Narrowing plausible paths to the target,
Organizing existing information into a more coherent structure,
Increasing confidence that the current reasoning context can be extended into a valid path reaching the target answer.

Reject the candidate if it only adds side branches or circular logic without clarifying the path to the target.

Output Constraints

Perform the reasoning progress implicitly (chain of thought).
Output ONLY one character: 1 or 0.
Output 1 ONLY if the candidate passes all three rules (i.e., it qualifies as a Gold Neighbor). Output 0 otherwise.

Question:
{question}

Target Answer:
{target_answer}

Current reasoning queue:
{queue_text}

Candidate Node:
{candidate_node_content}

Your answer (1/0 only):
\end{lstlisting}
\end{tcolorbox}

\vspace{1cm}

\begin{tcolorbox}[promptbox={Pairwise Preference Evaluation Prompt}]
\begin{lstlisting}
Task Description

You are comparing two candidate neighbors given the current reasoning queue.

The queue consists of text snippets and/or triples that have already been selected and should be treated as current working assumptions in the reasoning process. They define the current partial reasoning structure and direction, which may be incomplete or provisional, but must not be overturned without explicit support.

Select the candidate that, when added to this queue, most plausibly advances the reasoning toward answering the question.

Evaluation Rules (Apply IN ORDER, do NOT skip steps)

You must evaluate the two candidates strictly in the following order. If a candidate fails an earlier rule, it should NOT be preferred even if it looks good under later rules.

Execution Priority Summary

Rule 1 filters invalid candidates.
Rule 2 is the primary decision criterion.
Rule 3 is used only as a tie-breaker.
Output "0" only if no rule yields any preference.

Rule 1 - Context Consistency & Answer Reachability (Hard Gate)

Evaluate each candidate under the semantic and structural context defined by the current queue. The queue defines the current reasoning context and direction, which may be partial or provisional.

Candidates that rely on unsupported free association unrelated to the current context, or that arbitrarily overturn the current reasoning direction without justification, are forbidden.

Do not select candidates solely based on surface semantic similarity. Whether a candidate contributes meaningfully should be judged using the more detailed criteria in Rules 2 and 3.

Rule 2 - Missing Information Completion (Primary Criterion)

First determine:
what information is required to answer the question,
what is already present in the current queue,
and what specific information is still missing to complete a plausible answer path.

Missing information may include, but is not limited to:
A key entity or concept,
A necessary intermediate entity or role,
A relation, causal link, or dependency,
A factual constraint or evidence (e.g., authorship, time, location, identity, attribution),
A bridging fact that connects the current reasoning context to the final answer.

Such information may be introduced explicitly (via entities or relations) or implicitly (via text snippets that provide required facts or constraints).

Prefer candidates that directly fill one clearly identifiable missing component or make it clearly reachable from the current queue.

You should be able to clearly name the missing information, as a concrete component of the answer path, that the selected candidate helps to fill.

Rule 3 - Structural Progress (Secondary)

Apply this rule only if Rule 2 does not give a clear choice.

If only one candidate clearly fills missing information (Rule 2), select it directly.

If both or neither candidates fill missing information, compare how each candidate changes the structure of the reasoning when added to the current queue.

Prefer the candidate that makes the reasoning path toward the final answer more explicit, more directed, and less ambiguous, by:
Narrowing plausible answer paths,
Organizing existing information into a more coherent structure,
Increasing confidence that the current reasoning context can be extended into a valid answer path.

Do not reward candidates that only add side branches without clarifying the answer path.

Rule 4 - Redundancy Penalty (Fallback)

If none of the above rules distinguish the candidates, penalize candidates that only repeat or paraphrase information already present in the queue.

Output Constraints

Perform the reasoning progress implicitly(chain of thought).
Output ONLY one character: A, B, or 0.
You MUST select A or B if either candidate shows any identifiable advantage.
Output "0" only if neither candidate satisfies Rule 2 or Rule 3 in any identifiable way.
This should be rare.

Question:
{query}

Current reasoning queue:
{queue_text}

Candidates:
A: {pos_text}
B: {neg_text}

Your answer (A/B/0 only):
\end{lstlisting}
\end{tcolorbox}
\vspace{1cm}

\begin{tcolorbox}[promptbox={SCST Reward Evaluation Prompt}]
\begin{lstlisting}
You are a strict factual verifier. 
Question:
{question}
Given information:
{queue_text}
Correct answer:
{ground_truth}
Task:
Determine whether the given information is sufficient to confidently answer the question with the correct answer above.
Rules:
- Do NOT answer the question.
- Do NOT use any external knowledge.
- Only judge sufficiency of the given information.
- Reply with exactly one token: YES or NO.
\end{lstlisting}
\end{tcolorbox}

\end{document}